\documentclass[letterpaper, 10 pt, conference]{ieeeconf}  

\IEEEoverridecommandlockouts                              

\usepackage{mathtools}
\usepackage{gensymb}
\usepackage[export]{adjustbox}
\usepackage{color}
\usepackage{graphics} 
\usepackage{epsfig} 
\usepackage{times} 
\usepackage{amsmath} 
\usepackage{tabularx}
\usepackage{amssymb}  
\usepackage{leftidx}
\usepackage{tablefootnote}
\usepackage[pagebackref]{hyperref}
\usepackage{cite}
\hypersetup{
    colorlinks=true,
    citecolor=ForestGreen,
    linkcolor=blue,
    filecolor=magenta,      
    urlcolor=cyan
}
\usepackage{booktabs}
\usepackage{multirow}
\usepackage{multicol}
\usepackage{graphicx}
\usepackage{color}
\usepackage[font=small]{caption}
\usepackage{xfrac}
\usepackage{subcaption}     

\title{\LARGE \bf
GDIP: Gated Differentiable Image Processing for Object-Detection in Adverse Conditions
}

\author{Sanket Kalwar$^{*1}$, Dhruv Patel$^{*1}$, Aakash Aanegola$^{1}$, Krishna Reddy Konda$^{3}$,  Sourav Garg$^{2}$,  K Madhava Krishna$^{1}$
\thanks{* denotes equal contribution}
\thanks{$^{1}$are with RRC, IIIT Hyderabad, India {\tt\small \{sankethkalwar, dhruv.r.patel14, aakash.aanegola\}@gmail.com, mkrishna@iiit.ac.in}}%
\thanks{$^{2}$is with the QUT Centre for Robotics at the Queensland University of Technology (QUT), Brisbane, Australia.
        {\tt\small s.garg@qut.edu.au}}%
\thanks{$^{3}$is with ZF TCI, Hyderabad, India
        {\tt\small krishna.konda@zf.com}}
\thanks{$^\dagger$Project page: \href{https://gatedip.github.io}{https://gatedip.github.io} }
\thanks{$^\dagger$Code: \href{https://github.com/Gatedip/GDIP-Yolo}{https://github.com/Gatedip/GDIP-Yolo}}
}

\makeatletter
\let\@oldmaketitle\@maketitle
\renewcommand{\@maketitle}{\@oldmaketitle
\centering
\vspace{-1mm}

\includegraphics[width=\textwidth]{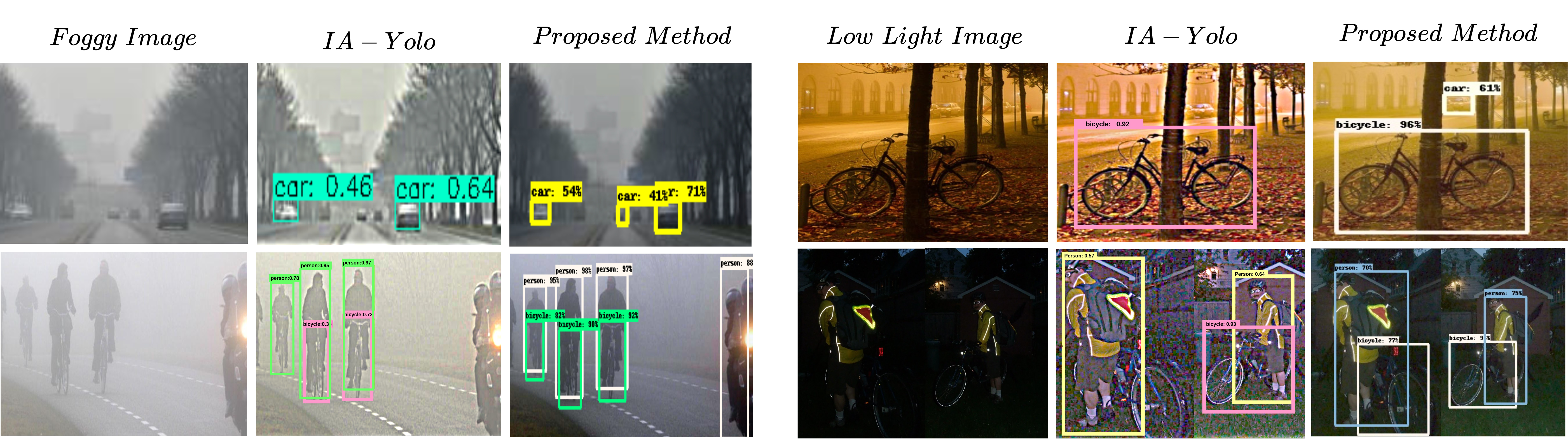}
\vspace{-5mm}
\captionof{figure}{\small{ \textbf{Overview}: Object detection is challenging in adverse weather conditions because objects in the scene are only partially visible, resulting in missed detections. We compare our proposed GDIP-Yolo with the current state-of-the-art (SOTA) IA-Yolo~\cite{liu2022imageadaptive} qualitatively. GDIP-Yolo detects more objects than IA-Yolo (the middle car (row 1, column 2) in the top foggy image, and the car in the background (row 1, column 5) in the top low-light image), and is more confident about its prediction (best viewed in 4x zoom). The keynote is GDIP-Yolo's informed weighting of differentiable Image Processing blocks that act concurrently on the input image, leading to a vastly superior detection performance than IA-Yolo's sequential image processing framework. Note: The confidence boxes for IA-Yolo are given as a fraction between [0,1], whereas ours is in percentage.} 
}
\label{fig:qualitative}
\vspace{0mm}
}
        
\begin{document}

\maketitle
\thispagestyle{empty}
\pagestyle{empty}


\begin{abstract}

Detecting objects under adverse weather and lighting conditions is crucial for the safe and continuous operation of an autonomous vehicle, and remains an unsolved problem.
We present a \underline{G}ated \underline{D}ifferentiable \underline{I}mage \underline{P}rocessing (GDIP) block, a domain-agnostic network architecture, which can be plugged into existing object detection networks (e.g., Yolo) and trained end-to-end with adverse condition images such as those captured under fog and low lighting. Our proposed GDIP block learns to enhance images directly through the downstream object detection loss. This is achieved by learning parameters of multiple image pre-processing (IP) techniques that operate \textit{concurrently}, with their outputs combined using weights learned through a novel \textit{gating mechanism}. We further improve GDIP through a multi-stage guidance procedure for progressive image enhancement. Finally, trading off accuracy for speed, we propose a variant of GDIP that can be used as a \textit{regularizer} for training Yolo, which eliminates the need for GDIP-based image enhancement during inference, resulting in higher throughput and plausible real-world deployment. We demonstrate significant improvement in detection performance over several state-of-the-art methods through quantitative and qualitative studies on synthetic datasets such as PascalVOC, and real-world foggy (RTTS) and low-lighting (ExDark) datasets. 
\end{abstract}
%
\section{INTRODUCTION}
Autonomous mobile agents need a high-level understanding of their environment to plan their trajectories and function effectively. This requires robust perception which stems from object detection and semantic segmentation like tasks for recognizing and localizing safety-critical objects such as pedestrians and vehicles. Most object detection methods are designed for and trained with images captured under ideal environmental conditions, and often do not generalize to adverse settings (like foggy and low-light conditions). Recent attempts to achieve such robustness include domain classification based invariant detection~\cite{chen2018domain,Hsu_2020_WACV,Tian_2021_ICCV,wu2021vector,wu2021instance,lin2021domain,gu2021pit,he2020domain}, prior-knowledge based feature adaptation~\cite{sindagi2020prior,zhang2020unified}, adversarially-trained image alignment~\cite{zhang2021domain,zhuang2020ifan}, map-specific domain adaptation~\cite{sakaridis2020map}, and physics-prior based zero-shot learning~\cite{lengyel2021zero, Zheng_2022_WACV}.

More recently, learnable image pre-processing methods have emerged as a superior alternative~\cite{liu2022imageadaptive,lengyel2021zero,huang2020dsnet,liu2019griddehazenet,guo2020zero,hu2018exposure,Zheng_2022_WACV,Yang_2020_CVPR,LI2021106617,Zhang_2020_ACCV}. However, these methods are either limited to a single preprocessing module~\cite{lengyel2021zero,Zheng_2022_WACV,Yang_2020_CVPR,LI2021106617,Zhang_2020_ACCV}, require domain-specific architectural variations~\cite{liu2019griddehazenet,guo2020zero,liu2022imageadaptive} or use multiple modules in an arbitrary sequential order~\cite{hu2018exposure,liu2022imageadaptive}. 
In this work, we address all these limitations and present a novel approach that learns to enhance images for object detection under adverse conditions in an end-to-end  manner. This is achieved through a learnable gating-based weighted combination of concurrent image processing operations, dubbed GDIP (Gated Differentiable Image Processing). 
Our proposed GDIP method integrated with Yolo significantly outperforms the current SOTA, Image Adaptive Yolo (IA-Yolo~\cite{liu2022imageadaptive}), which relies on an arbitrary sequential image preprocessing. On real-word fog (RTTS~\cite{RTTS}) and low-light (Ex-Dark~\cite{Exdark}) datasets, GDIP-Yolo leads IA-Yolo in mAP by 5.76 and 15.89 respectively. The key contributions of this paper are listed below:
\begin{enumerate}
    \item a novel \textit{gating mechanism} that enables \textit{concurrent} relative weighting of multiple differentiable image processing modules to enhance images for object detection under adverse environmental conditions;
    \item a \textit{multi-level} version of GDIP where an image is progressively enhanced through multiple GDIP blocks, each guided by a different layer of the image encoder; and
    \item an adaptation of GDIP as a \textit{training regularizer} which directly improves object detection training for adverse conditions, eliminating the need of GDIP during inference, thus saving compute time with a minor drop in performance.
\end{enumerate}

\section{RELATED WORK}
\label{sec:related_work}
Object detection is the problem of localizing and classifying objects in the scene and has seen a recent uptick in popularity due to its applications in autonomous vehicles and more. There are two primary approaches to the object detection problem, two-stage detection, and single-stage detection. Two-stage detectors like FasterRCNN~\cite{ren2015faster} and MaskRCNN~\cite{he2017mask} utilize a region proposal network (RPN) which generates proposals of plausible regions of interest, sent to the downstream network that performs classification. Two-stage approaches are computationally expensive, reducing their application range. Single-stage detectors like Yolo~\cite{redmon2018yolov3}, RetinaNet~\cite{lin2017focal}, SSD~\cite{liu2016ssd} and FCOS~\cite{tian2019fcos} bypass the heavy RPN and directly extract objects with their associated labels. 
Nevertheless, under adverse conditions, both types of networks fail to detect objects.
\textbf{Adverse Conditions:} Typical object detection techniques fail in adverse weather conditions, and transfer learning has proven to be a viable way to employ object detection in adverse weather conditions. Chen et al.~\cite{chen2018domain} approach this problem from a domain adaptation perspective and utilize image and instance-level features to reduce the domain shift. Singadi et al.~\cite{sindagi2020prior} use weather-specific knowledge and define a prior-adversarial loss with feature recovery to mitigate weather effects on detection. Multiscale Domain Adaptive Yolo (MS-DAYOLO)~\cite{hnewa2021multiscale} uses classifiers for each domain at different scales to learn domain invariant features. Zhang et al.~\cite{zhang2021domain} employ image-level feature alignment to match local and global features. 

\textbf{Differentiable Image Pre-processing:} Another popular approach to the problem is to perform image enhancement before object detection. In Exposure~\cite{hu2018exposure}, a deep Reinforcement Learning model learns a policy to apply a sequence of enhancement operations. AOD-Net~\cite{li2017aod} dehazes images using a CNN designed on a re-formulated atmospheric scattering model.
Dong et al.~\cite{dong2020multi} use an encoder-decoder architecture (U-Net) with the strength-operate-subtract boosting strategy to help dehaze images. GridDehazeNet~\cite{liu2019griddehazenet} employs a multi-scale attention mechanism with pre and post-processing modules to generate better inputs and reduce artifacts in the final dehazed image. He et al.~\cite{he2010single} use a dark channel prior (one color channel in most pixels will be low) to dehaze images but do not perform object detection. Some models target only a single adverse condition, Guo et al.~\cite{guo2020zero} perform light enhancement by estimating light enhancement curves that are applied iteratively to the input image, boosting face detection performance. Zeng et al.~\cite{zeng2020learning} learn multiple look-up tables and use CNN predictions to fuse the look-up tables into one and transform the color and tone of a source image. DSNet~\cite{huang2020dsnet} uses two subnets for image restoration and object detection to boost performance in adverse weather conditions. IA-Yolo~\cite{liu2022imageadaptive} uses a CNN to predict differentiable image processing parameters trained in conjunction with Yolov3~\cite{redmon2018yolov3} for object detection to enhance images, and perform object detection in an end-to-end fashion.

Unlike existing methods, GDIP is a domain agnostic network architecture that handles multiple image processing operations concurrently. Additionally, it has a unique advantage with its utility as a training regularizer, which eliminates image enhancement overhead during inference resulting in higher throughput. 


\section{PROPOSED METHOD}
\label{sec:proposed_method}
We propose a Gated Differentiable Image Processing (GDIP) framework that learns to enhance input images for object detection in adverse environmental conditions. The GDIP block learns parameters for multiple Image Processing (IP) operations performed concurrently and learns the optimal weights to combine their output.  
We use the following IP operations (similar to
IA-Yolo~\cite{liu2022imageadaptive}): tone correction ($T$), contrast balance ($C$), sharpening ($S$), defogging ($DF$), gamma correction ($G$), white balancing ($WB$), and the identity operation ($I$).
Unlike IA-Yolo's sequential image enhancement, GDIP enhances images through a weighted combination of concurrent IP operations.\looseness=-1

\subsection{\textbf{Gated Differentiable Image Processing (GDIP) block:}}
\label{sub:GDIP-block}

\renewcommand{\thefigure}{2}
\begin{figure}
\includegraphics[scale=0.3]{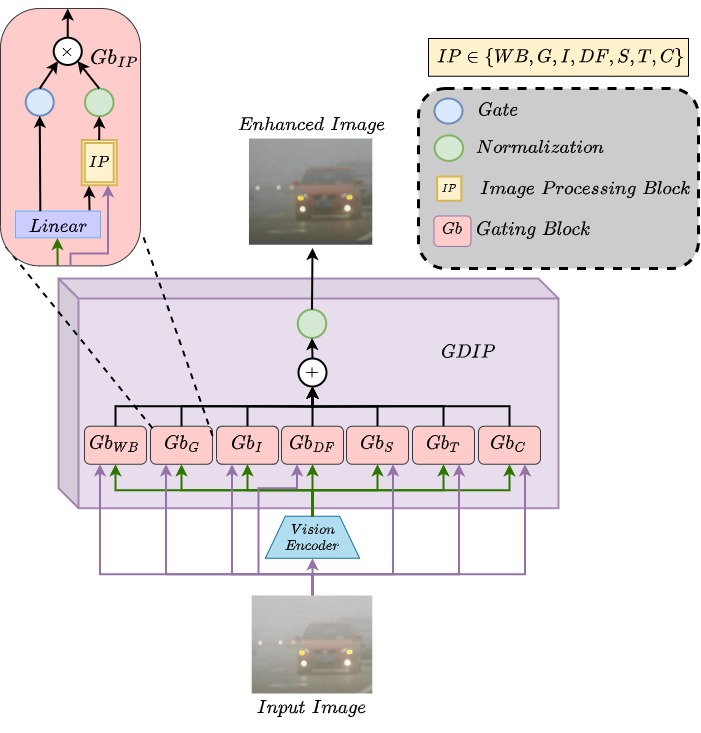}
\centering
\caption{\textbf{GDIP Block:} The peach blocks represent different $Gb_{IP}$ and their structures can be seen in the expanded general $Gb$ block (top left corner). Yolo and the object detection output are not shown due to space constraints.}
\label{fig:singel-level-gdip}
\end{figure}

The GDIP block (shown in Fig.~\ref{fig:singel-level-gdip}) consists of multiple gated image processing modules, referred to as $Gb_{IP}$, that individually enhance images, which are then combined through the weights predicted by the gates. Each $Gb$ module contains a linear layer, a differentiable image processing operation, a gate (shifted \textit{tanh} function that returns a value between 0 and 1), and a normalization operation. The linear layer (purple linear block in Fig.~\ref{fig:singel-level-gdip}) computes two entities: the parameters required by the differentiable IP block and a scalar value that serves as an input to its corresponding gate. The individual linear layers of every $Gb$ module are passed a common feature embedding as input, obtained from a shared vision encoder (described later). The output of the IP operation (using the predicted parameters) gets multiplied by the scalar output of the gate. The weighted outputs of individual $Gb$ blocks are finally aggregated to obtain an enhanced image. Expressed mathematically, the output of the GDIP block is: 
\begin{equation}
z = N(\sum_{i}N(f_{i}(x)) * w_{i})
\label{eqn:gdip_main}
\end{equation}
where $x$ is the input image captured under adverse environmental conditions, $z$ is the enhanced clear image, $f_{i}(x)$ represents the $i^{th}$ IP operation (top-right in Fig.~\ref{fig:singel-level-gdip}) weighted by its respective scalar gate output $w_{i} \in [0,1]$, and $N$ is the min-max normalization operation. Normalization ensures that the pixel intensity range of the output of all the image processing operations are the same. 
The IP operations are expressed mathematically in Table~\ref{tab:IP_operations}, see~\cite{liu2022imageadaptive} for a detailed description. 

\begin{table}
\scriptsize
\caption{Image Processing (IP) Operations. The parameters in the second column are computed by the linear layer in GDIP. Further details on the equations can be found in IA-Yolo~\cite{liu2022imageadaptive}}
\centering  
\begin{tabular}{l c  c}  
\toprule
\\ [-2ex]
IP Operation
&  Parameters & Function \\ [0.5ex]
\midrule
Tone & $t_{i}$ & \hspace{-7mm} $I_{tone} = (L_{t_{r}}(r_{i}),L_{t_{g}}(g_{i}),L_{t_{b}}(b_{i}))$\\ [0.5ex]
Contrast & $\alpha$ & $I_{contrast}=\alpha*En(I) + (1-\alpha)*I$ \\ [0.5ex]
Sharpening & $\lambda$ & \hspace{-6mm}$I_{sharpen} = I + \lambda*(I - Gaussian(I))$ \\ [0.5ex]
Defogging & $\omega$ & \hspace{-5mm}$I = I_{defog}*t(\omega)+A*(1-t(\omega))$ \\ [0.5ex]
Gamma & $\gamma$ & $I_{gamma} = I^{\gamma}$ \\ [0.5ex]
White balance & $W_{r},W_{g},W_{b}$ & $I_{wb} =(W_{r}r_{i},W_{g}g_{i},W_{b}b_{i})$ \\[0.5ex]
Identity & - & $I_{identity} = I$ \\[0.5ex]
\bottomrule
\end{tabular}
\label{tab:IP_operations}
\end{table}
\noindent\textbf{Vision Encoder:}
Our proposed GDIP block requires latent embeddings to compute image processing parameters and gate values. For this purpose, we employ a vision encoder comprising five convolutional layers (each with a kernel size of three and a stride of one). The number of channels in each layer is double the previous, starting from 64 in the first layer and 1024 in the final layer. Each convolution operation is followed by average pooling (with kernel size three and stride two), while the last layer is followed by global average pooling, the output of which is a 1x1x1024. This is then projected to a 256-dimensional latent space using a fully connected layer. The GDIP block takes this 256-dimensional embedding from the vision encoder along with the adverse input image and performs image enhancement after computing the necessary parameters.

\noindent\textbf{GDIP-Yolo:}
To integrate GDIP with Yolo, we use the vision encoder with GDIP to perform image enhancements (depicted in Fig.~\ref{fig:singel-level-gdip}), and use the enhanced image as input to Yolo. Integrating GDIP with Yolo in this fashion ensures that our architecture doesn't require any additional loss formulation and uses Yolo's standard object detection loss~\cite{redmon2016you} (referred to as $L_{obj}$) to train the network for object detection end-to-end. 




\renewcommand{\thefigure}{3}
\begin{figure}
\includegraphics[width=0.5\textwidth,height=0.16\textwidth]{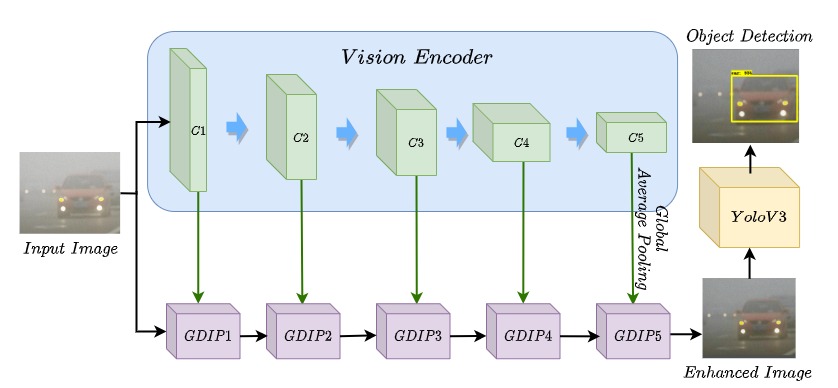}
\centering
\caption{\textbf{MGDIP-Yolo:} The GDIP blocks perform enhancements
progressively on the input image starting with $GDIP_1$ obtaining
intermediate features from $C1$ (each GDIP block still performs
IP operations concurrently). The final enhanced image is passed to
Yolo and the whole pipeline is trained end-to-end.}
\label{fig:mgdip-yolo}
\end{figure}
\label{sub:MGDIP-Yolo}

\subsection{\textbf{Multi-Level GDIP (MGDIP):}}
GDIP-Yolo contains a single GDIP block, which is fed with latent embeddings obtained from the vision encoder. Since, we only use the last layer of the vision encoder for this purpose, it limits the extent of information available for GDIP to learn parameters for image processing modules. Thus, we propose multi-level progressive image enhancement, achieved by integrating a GDIP block with every layer of the vision encoder, dubbed MGDIP-Yolo. Note that the individual image processing modules within a single GDIP block still operate concurrently with their corresponding gates providing relative weightings. As shown in Fig.~\ref{fig:mgdip-yolo}, MGDIP progressively enhances images by feeding the output from one GDIP block as input to the next, where individual GDIP blocks are guided by the features extracted from different layers of the vision encoder. The final enhanced output from MGDIP is passed to Yolo for object detection. MGDIP-Yolo is trained in an end-to-end manner using the standard object detection loss $L_{obj}$, similar to GDIP-Yolo.

We hypothesize that utilizing embeddings from different layers provides GDIP access to multiple feature scales, each of which can have a varied relevance for different image processing operations.
This is based on the understanding that earlier layers in CNNs capture lower level information (local information like edges) and later layers capture high-level (global) information. 
Thus, MGDIP gains the ability to use the local/global feature properties to selectively apply image processing operations.

\renewcommand{\thefigure}{4}
\begin{figure}
\includegraphics[scale=0.3]{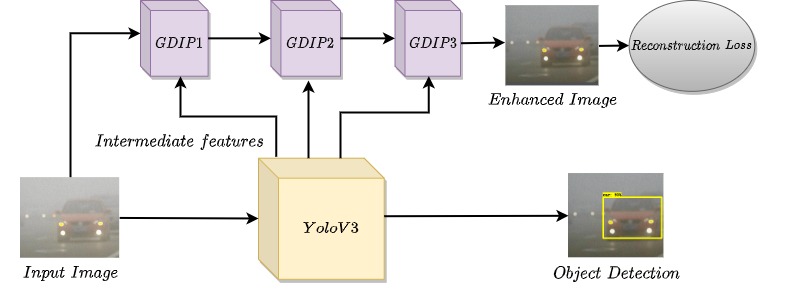}
\centering
\caption{\textbf{GDIP as a regularizer:} The purple GDIP blocks and the reconstruction loss help regularize Yolo's features. During inference, everything apart from Yolo is removed.}
\label{fig:GDIP-reg}
\end{figure}

\label{sub:GDIP-reg}
\subsection{\textbf{GDIP block as a regularizer:}}
In this section, we demonstrate how GDIP can also be employed as a feature regularization technique to improve Yolo's performance while maintaining its throughput. 

The original GDIP block used a vision encoder to obtain feature embeddings. Alternatively, multiple GDIP blocks can be connected to intermediate layers of Yolo, bypassing the need for a vision encoder and directly using Yolo's embeddings to construct an enhanced output, as shown in Fig.~\ref{fig:GDIP-reg}. Note that this enhanced output is not the input to Yolo but rather a byproduct that we use for training regularization. The reconstruction loss (Eq.~\ref{eqn:rec_loss}) is calculated between this output and the clear version of the input image as a combination of $L_{1}$ norm and Mean Square Error loss $L_{MSE}$. The overall loss function used is shown in Eq.~\ref{eqn:total_regularizer_loss}, where $\alpha$ is the weight of the reconstruction loss and is empirically set to $1\times10^{-4}$. 
\noindent \begin{minipage}[b]{.49\linewidth}
\begin{equation}
 \small
 L_{Reg} = L_{1} + L_{MSE}  \label{eqn:rec_loss}
\end{equation}
\end{minipage}
\begin{minipage}[b]{.49\linewidth}
\begin{equation}
 \small
L_{total} = L_{obj} + \alpha L_{Reg} \label{eqn:total_regularizer_loss}
\end{equation}
\end{minipage}


Inclusion of the reconstruction loss in the formulation helps Yolo learn features that are invariant to adverse conditions, resulting in better performance when compared to standalone Yolo. Since the GDIP blocks exist solely to refine Yolo's features, it is only required during training and can be removed during inference. This results in an unchanged network architecture (Yolo) that performs better in adverse weather conditions along with higher throughput.

\section{EXPERIMENTAL SETUP}
\label{sec:exp_setup}

\subsection{Datasets}
\noindent\textbf{Foggy Conditions:} 
We use the RTTS dataset~\cite{RTTS}, a collection of 4322 natural foggy images with five annotated classes - person, car, bus, bicycle and motorcycle - primarily for testing. The PascalVOC train/val datasets (2007 and 2012)~\cite{pascal-voc-2007,pascal-voc-2012} have 22136 clear images that we use as a base to create a synthetic training set. We select images from PascalVOC having objects belonging to the five classes from RTTS and create two datasets, one with clear images (VOCNormal) and one with augmented foggy images generated using the atmospheric scattering model (ASM)~\cite{AtmospshereScatteringModel}. We employ the ASM to generate 10 different levels of fog to include variance in our synthetic training set. We subsample and prepare a synthetic testing set in a similar fashion from the PascalVOC 2007 test set with 4952 images (referred to as V\_F\_Ts). We employ a \textit{hybrid} strategy where we use a mix of foggy and clear images (in a 2:1 ratio) to help our model learn fog-invariant features. 

\noindent\textbf{Low-lighting Conditions:} 
The ExDark dataset~\cite{Exdark} is a collection of 7363 real-world images with 10 object classes in low lighting conditions that we use to evaluate our models. Similar to preparing the foggy dataset, we select images from PascalVOC having objects from the 10 classes of ExDark and apply a gamma filter to emulate a low-lighting condition. Mathematically, $I_{dark} = I^{\gamma}$, where $\gamma$ is sampled uniformly from the range of 1.5 to 5, $I$ is the normalized clear image, and $I_{dark}$ is the synthetic dark image. Using the same selection and image processing methods, we generate a synthetic low-light test set (V\_D\_Ts) from the PascalVOC test set. During training, we employ a \textit{hybrid} strategy (similar to the foggy setting) by using a mix of dark and clear images.


\subsection{Training Setup}
Training for both foggy and low-lighting setting is done by resizing images to $448 \times 448 \times 3$ pixels and with a batch size of 6 for 80 epochs. We use a cosine learning rate scheduler with learning rates ranging from $1\times 10^{-6}$ to $1\times 10^{-4}$ and an SGD optimizer with a weight decay of $5\times 10^{-4}$.

\section{RESULTS AND ANALYSES}
\label{sec:results_analyses}

We provide qualitative and quantitative results that establish our proposed method's superiority and evaluate design choices through ablation studies. We also show that GDIP variants provide flexibility to prioritize speed or accuracy based on application requirements.

\subsection{Qualitative Analysis}
We compare the results of GDIP-Yolo with the current SOTA IA-Yolo, shown in Fig.~\ref{fig:qualitative} on real-world data. Unlike IA-Yolo, our method clears fog and improves lighting conditions without changing underlying color distributions. Our method is able to detect far-off objects such as cars and bicycles (see the third column in Fig.~\ref{fig:qualitative}), which are generally missed in extreme foggy conditions by SOTA IA-Yolo (see the second column). The last column (low-light conditions) clearly indicates GDIP-Yolo enhances lighting without artifacts, helping the model detect all objects in the scene (the car in the background, for example). We quantify the improvement extended by GDIP in the next section. 

\subsection{Quantitative Analysis}
We compare our proposed method with other SOTA works using the standard object detection evaluation metric - mean average precision (mAP). All mAP values are calculated at an IoU (Intersection over Union) of 0.5.

\paragraph{Foggy Conditions} In Table~\ref{tab:quantitative_foggy}, we compare our proposed variants of the GDIP with other competing methods on VOCNormal Test set (V\_N\_Ts), synthetic VOCFoggy Test set (V\_F\_Ts), and the real-world foggy dataset RTTS. The second column in the table shows the training data used by each of the methods, where ``Hybrid'' implies the use of both clear and foggy data. We set YoloV3 as the baseline, which is trained on a mix of foggy and clear images to validate if we can improve performance by using data augmentation. We also compare our results against a diverse range of methods based on domain adaption (DA-Yolo~\cite{hnewa2021multiscale}), multi-task learning (DSNet~\cite{huang2020dsnet}), defogging as pre-processing (MSBDN~\cite{dong2020multi}, GridDehaze~\cite{liu2019griddehazenet}), and adaptive image enhancement (IA-Yolo~\cite{liu2022imageadaptive}).
It can be observed in Table~\ref{tab:quantitative_foggy} that our proposed variants of GDIP establish a new SOTA across different fog datasets.

\begin{table}[]
\scriptsize
\caption{Quantitative results for \textit{foggy} conditions on the {V\_N\_Ts} (VOCNormal Test set), {V\_F\_Ts} (VOCFoggy Test set) and real-world RTTS dataset. Best and second best mAP scores are bold and italicized, respectively. }
\centering  
\begin{tabular}{l l l c  r r}  
\hline\hline                       
\\ [-2ex]
\multicolumn{2}{c}{\raisebox{-1.5ex}{Methods}} & \raisebox{-1.5ex}{Train Data} &  V\_N\_Ts & V\_F\_Ts & RTTS\\ [-1.0ex]
$~$ & $~$ & & (mAP) & (mAP) & (mAP) \\ [0.5ex]
\hline \hline 
\\ [-2ex]

Baseline & Yolov3~\cite{redmon2018yolov3} & Hybrid & 64.13 & 63.40 & 30.80 \\ [0.5ex]
\hline \\[-1.5ex]

& MSBDN~\cite{dong2020multi} & {\scriptsize VOC\_Norm} & - & 57.38 & 30.20  \\ [-1ex]
\raisebox{1.5ex}{Defog} & GridDehaze~\cite{liu2019griddehazenet} &  {\scriptsize VOC\_Norm} & - & 58.23 & 31.42  \\ [0.5ex]
\hline \\[-1.5ex]

Domain & DAYolo~\cite{hnewa2021multiscale} & Hybrid & 56.51 & 55.11 & 29.93 \\  [0.5ex]
Adaptation & & &\\ [0.5ex]
\hline \\[-1.5ex]

Multi-task & DSNet~\cite{huang2020dsnet} & Hybrid & 53.29 & 67.40 & 28.91 \\ [0.5ex]
\hline \\[-1.5ex]

Image & IA-Yolo~\cite{liu2022imageadaptive} & Hybrid & 73.23 & 72.03 & 37.08 \\ [0.5ex]
Adaptive & & &\\ [0.5ex]
\hline \\[-1.5ex]

 
 
 & GDIP-Yolo & Hybrid & \textit{73.70} & 71.92 & \textit{42.42} \\ [0.25ex]

{Proposed}
  &\raisebox{-.65ex}{MGDIP-Yolo} &\raisebox{-.5ex}{Hybrid} & \raisebox{-.5ex}{\textbf{75.36}} & \raisebox{-.5ex}{\textbf{73.37}} & \raisebox{-.5ex}{\textbf{42.84}} \\ [0.25ex]
 {Method}

 
   &\raisebox{-.65ex}{GDIP} &\raisebox{-.5ex}{Hybrid} & \raisebox{-.5ex}{73.17} & \raisebox{-.5ex}{\textit{72.77}} & \raisebox{-.5ex}{39.52} \\ [-1ex]
 &\raisebox{-.65ex}{Regularizer} &  & & \\  [0.75ex]
 
\hline
\\ [-2ex]
\end{tabular}
\label{tab:quantitative_foggy}
\end{table}

All GDIP variants  perform significantly better than SOTA methods on RTTS, which tests the generalizability of our method to real-world conditions. Our basic GDIP-Yolo variant outperforms the SOTA method IA-Yolo by 5.34 mAP. This can be attributed to the concurrently weighted IP operations unlike IA-Yolo's fixed sequential pipeline. MGDIP-Yolo further improves upon the GDIP-Yolo by 0.42 mAP and does so consistently across all datasets. It emerges superior to all other methods and GDIP variants, as it benefits from multi-scale information. Our regularizer variant outperforms all SOTA methods, while being as fast as vanilla Yolo (see Subsection \ref{subsec:real-time}), emerging as an alternative to GDIP-Yolo and MGDIP-Yolo with an accuracy-speed trade-off. 

\paragraph{Low-lighting conditions}
We compare our proposed variants with other SOTA methods on the real-world ExDark dataset, synthetic low-lighting VOCDark test set (V\_D\_Ts), and VOCNormal test set (V\_N\_Ts), as shown in Table~\ref{tab:quantitative_dark}. Here once again, we set YoloV3 as the baseline trained on hybrid data of a mix of dark and clear images. In addition, we also compare against a diverse range of methods based on light enhancement as pre-processing (ZeroDCE~\cite{guo2020zero}), domain adaptation (DA-Yolo~\cite{hnewa2021multiscale}), multi-task learning (DSNet~\cite{huang2020dsnet}) and adaptive image enhancement (IA-Yolo~\cite{liu2022imageadaptive}).
\begin{table}[]
\scriptsize
\caption{Quantitative results for \textit{low-lighting} conditions on the {V\_N\_Ts} (VOCNormal Test set), {V\_D\_Ts} (VOCDark Test set) and real-world ExDark dataset. Best and second best mAP scores are bold and italicized, respectively.}
\centering  
\begin{tabular}{l l l c  r r}  
\hline\hline                       
\\ [-2ex]
\multicolumn{2}{c}{\raisebox{-1.5ex}{Methods}} & \raisebox{-1.5ex}{Train Data} &  V\_N\_Ts & V\_D\_Ts & ExDark\\ [-1.0ex]
$~$ & $~$ & & (mAP) & (mAP) & (mAP) \\ [0.5ex]
\hline \hline 
\\ [-2ex]

Baseline & Yolov3~\cite{redmon2018yolov3} & Hybrid & 62.73 & 52.28 & 37.03 \\ [0.5ex]
\hline \\[-1.5ex]

Enhance & ZeroDCE~\cite{guo2020zero} & {\scriptsize VOC\_Norm} & - & 33.57 & 34.41 \\  [0.5ex]
\hline \\[-1.5ex]

Domain & DAYolo~\cite{hnewa2021multiscale} & Hybrid & 41.68 & 21.53 & 18.15 \\  [0.5ex]
Adaptation & & &\\ [0.5ex]
\hline \\[-1.5ex]

Multi-task & DSNet~\cite{huang2020dsnet} & Hybrid & \textbf{64.08} & 43.75 & 36.97 \\ [0.5ex]
\hline \\[-1.5ex]

Image & IA-Yolo~\cite{liu2022imageadaptive} & Hybrid & 56.01 & 48.44 & 26.67 \\ [0.5ex]
Adaptive & & &\\ [0.5ex]
\hline \\[-1.5ex]
 
 
 & GDIP-Yolo & Hybrid & \textit{63.23} & \textit{57.85} & \textbf{42.56} \\ [0.15ex]
{Proposed}
  &\raisebox{-.65ex}{MGDIP-Yolo} &\raisebox{-.5ex}{Hybrid} & \raisebox{-.5ex}{62.86} & \raisebox{-.5ex}{\textbf{57.91}} & \raisebox{-.5ex}{\textit{40.96}} \\ [0.35ex]
 
 {Method} 
 
   &\raisebox{-.65ex}{GDIP} &\raisebox{-.5ex}{Hybrid} & \raisebox{-.5ex}{62.30} & \raisebox{-.5ex}{57.67} & \raisebox{-.5ex}{40.72} \\ [-1.25ex]
 &\raisebox{-.95ex}{Regularizer} &  & & \\  [0.85ex]
 
\hline
\\ [-2ex]
\end{tabular}
\label{tab:quantitative_dark}
\end{table}
Our proposed GDIP-Yolo outperforms all the existing methods on the real-world ExDark dataset and achieves an absolute increase of 16 mAP over the previous SOTA IA-Yolo. Additionally, MGDIP-Yolo and GDIP as a regularizer variants also perform superior to other methods in this setting. In the synthetic low-lighting setting, MGDIP-Yolo approach emerges superior, while the other GDIP variants also significantly improve the performance over other methods. On the VOCNormal test set, our proposed method performs comparable to DSNet~\cite{huang2020dsnet}. To conclude, our proposed variants are significantly better than existing approaches for low-light settings (synthetic and real-world).
\renewcommand{\thefigure}{5}
\begin{figure}[b]
\includegraphics[width=\columnwidth,height=0.5\columnwidth]{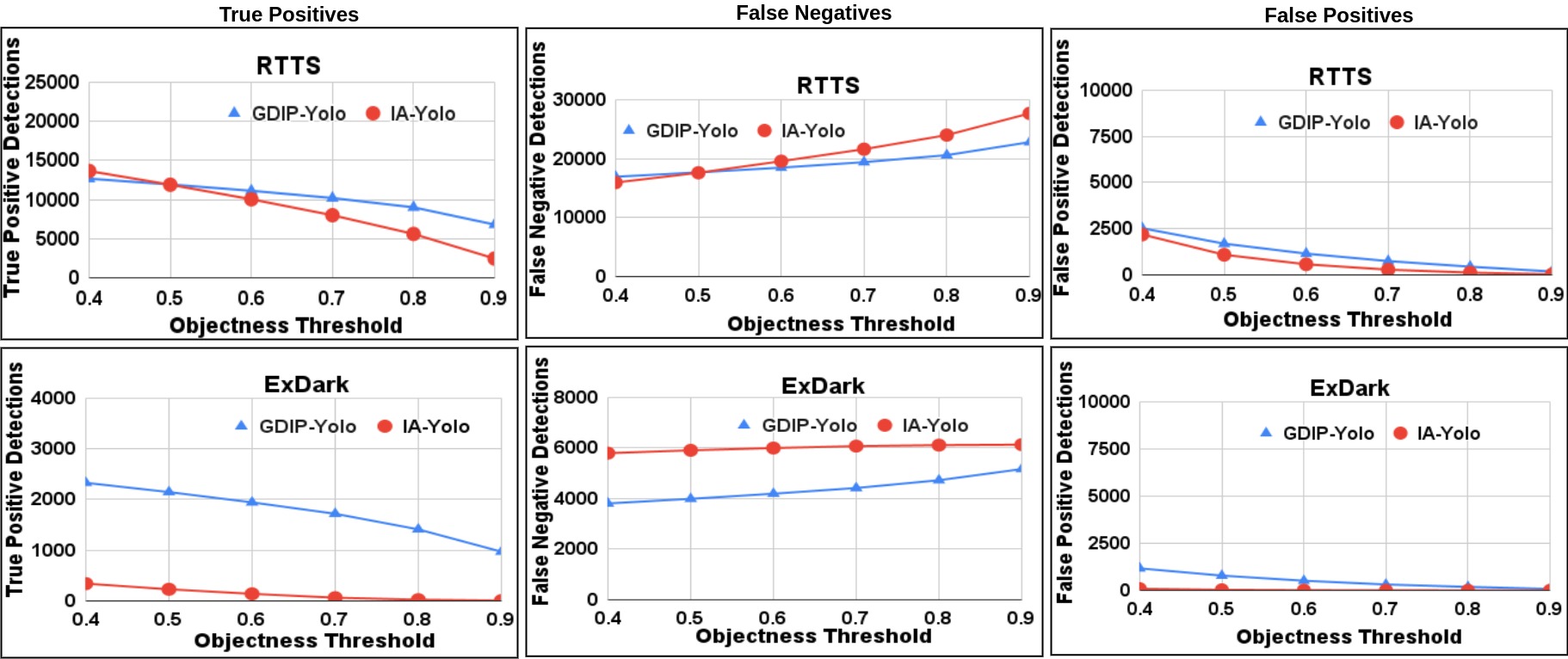}
\centering
\caption{\textbf{Alternative Metrics:} We indicate the True Positives (TP), False Negatives (FN) and False Positives (FP) on the RTTS and ExDark datasets for GDIP-Yolo (blue) and IA-Yolo (red). The x and y axes in the plots represent the objectness (detection confidence) threshold and the absolute number of TP/FN/FP. Detection confidence reflects the likelihood of box containing the object.
}
\label{fig:analysis_plot}
\end{figure}

\subsection{Detection Statistics}
 We present True and False Positives (TP, FP) and False Negatives (FN) of the number of object detections as an interpretable statistical measure as mAP does not convey the actual detections. The TP (Fig.~\ref{fig:analysis_plot} left) and FN (Fig.~\ref{fig:analysis_plot} middle) plots show substantial improvement at high object detection confidence thresholds both for RTTS and Ex-Dark datasets for GDIP-Yolo vis a vis SOTA IA-Yolo. 
For Autonomous Driving applications, the TP and FN statistics are critical, as not detecting an object when present can be catastrophic, and on these vital statistics, the significantly superior performance of GDIP is evident. GDIP-Yolo evaluates to comparable FP metrics vis a vis SOTA at high confidence thresholds on which it shows vastly superior performance on the TP, and FN metrics.

\subsection{Ablation Studies}

In this section, we demonstrate experimental support for using the gating mechanism and normalization layer in our proposed GDIP block. We validate that incorporating gating and normalization in the GDIP block provides a stable improvement in the downstream detection task for both foggy and low-light conditions (as shown in Table~\ref{tab:ablation_study}).

\renewcommand{\thefigure}{6}
\begin{figure}
\includegraphics[scale=0.2]{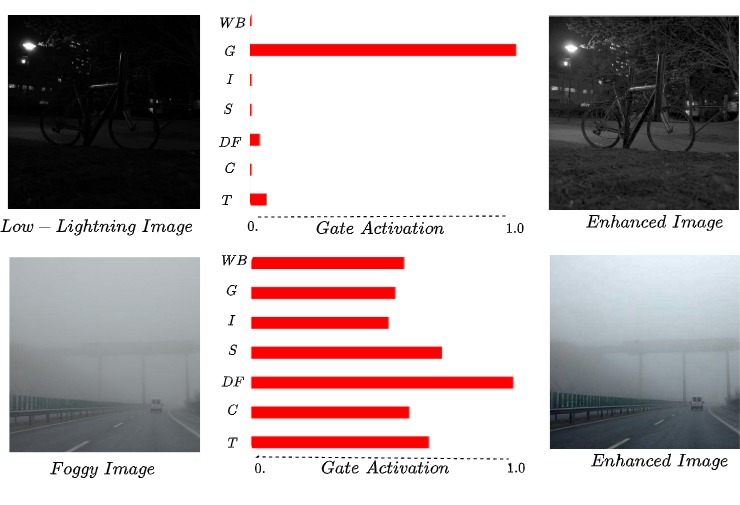}
\centering
\caption{\textbf{GDIP gate firing pattern:} The bar charts show the gates firing for low lighting and foggy images. GDIP learns the optimum enhancements to apply based on the input image features.}
\label{fig:GDIP-gatefiring}
\end{figure}

\noindent\textbf{Single Best vs Weighted Combination:}
Our proposed gating mechanism helps combine image processing operations through relative weighting (see Eq.~\ref{eqn:gdip_main}). To illustrate its effect, we remove this mechanism and use a single best image processing operation based on the highest gate value (referred to as GDIP-max in Table~\ref{tab:ablation_study}), expressed as $z = N(f_{i^*}(x))$ where $i^{*} = \mathrm{argmax}_i\ w_{i}$. Without the proposed gating, performance  reduces by 10.4 mAP for the RTTS dataset (comparing row 1 and 2 in Table~\ref{tab:ablation_study}). 
For the ExDark dataset the performance increases by a negligible amount - 0.15 mAP, insubstantial compared to the drop observed in the foggy setting. This study indicates that incorporating enhancements from multiple image processing operations is necessary as no single operation is  sufficient for dealing with adverse conditions. 


\noindent\textbf{Uniform vs Predicted Weighting:}
In this experiment, we compare our proposed relative weighting of image processing operations with a uniform weighting across all operations (referred to as GDIP w/o gates in Table~\ref{tab:ablation_study}), expressed mathematically as $z = N(\sum_{i}N(f_i(x)))$. We observe that mAP reduces by 0.77 and 0.27 for RTTS and ExDark, respectively. 
This performance drop can be attributed to the fact that depending on the environmental conditions of images, image processing operations need to be weighted differently, achieved through gating. This is evident from Fig.~\ref{fig:GDIP-gatefiring}, where activation of the gamma gate (G) is prominent for low light conditions, and in the case of foggy conditions, most of the gates remain activated with defogging (DF) being the highest. The gate activation patterns also help improve the interpretability of GDIP by indicating which IP operations are performed by what proportions based on the input image.

\noindent\textbf{With and Without Normalization:}
We also verify the necessity for the normalization layer after each image processing operation by removing them (GDIP-unnormalized in Table~\ref{tab:ablation_study}), expressed as $z = N(\sum_{i} f_{i}(x) * w_{i})$. This leads to a considerable performance drop of around 1.8 and 2.35 mAP in RTTS and ExDark, respectively. 

Overall, these ablation studies indicate that GDIP with normalization and the gating mechanism leads to the best overall performance irrespective of environmental conditions and is a promising solution for the object detection task. 



\begin{table}
\caption{Ablation Study}
\centering  
\resizebox{.3\textwidth}{!}{
\begin{tabular}{l c  c}  
\toprule
\textbf{Method}
&  \textbf{RTTS} (mAP) & \textbf{ExDark} (mAP) \\ [0.5ex]
\midrule
\\ [-2.0ex]
GDIP & $\mathbf{42.42}$ & 42.56 \\ [0.5ex]
GDIP-max & 31.99 & $\mathbf{42.71}$ \\ [0.5ex]
GDIP-unnormalized & 40.61 & 40.2 \\ [0.5ex]
GDIP w/o gates & 41.65 & 42.29 \\[0.5ex]
\bottomrule
\end{tabular}
\label{tab:ablation_study}
}
\end{table}

\subsection{Real-time performance}
\label{subsec:real-time}
In Table~\ref{tab:real_time}, we  compare the real-time performance of our proposed GDIP variants with other techniques. 
GDIP peforms the fastest as a regularizer at around 68 fps on a Nvidia GTX 1080Ti, which is the same as YoloV3. Our basic variant - GDIP-Yolo operates at 7 fps higher than IA-Yolo, while achieving SOTA mAP on real-world fog and night datasets.

\begin{table}
\caption{Real-time performance on GeForce GTX 1080}
\centering  
\resizebox{.3\textwidth}{!}{
\begin{tabular}{l c }  
\toprule
\textbf{Methods}
&  \textbf{FPS}  \\ [0.5ex]
\midrule

Yolo V3 & 68.39 $\pm$ 1.5 \\ [0.5ex]
\midrule

IA-Yolo  & 22.84 $\pm$ 0.0\\ [0.5ex]
\midrule

GDIP-Yolo & 29.78 $\pm 0.1$ \\ [0.5ex]
MGDIP-Yolo (top-down) & 11.38 $\pm$ 0.02\\ [0.5ex]
MGDIP-Yolo (bottom-up)  & 11.25 $\pm$ 0.03\\ [0.5ex]
GDIP as regularizer  & \textbf{68.39} $\pm$ \textbf{1.5}\\ [0.5ex]

\bottomrule
\end{tabular}
\label{tab:real_time}
}
\end{table}

\section{CONCLUSION}
\label{sec:conclusion}
We presented GDIP and MGDIP as domain-agnostic network architectures for object detection in adverse weather conditions, which can be used with existing object detection networks and trained under different adverse conditions, as we demonstrated for fog and low lighting. We also presented a training regularizer variant of GDIP, which improves the baseline Yolo performance under adverse conditions while maintaining its original throughput. All our GDIP variants result in a new state-of-the-art on challenging real-world datasets both under foggy and low-lighting conditions, while only having trained on synthetic adverse condition data, thus exhibiting significant generalization capability.

In future, this work can be extended to other adverse condition types (e.g., haze, rain, snow, etc.) along with additional relevant image pre-processing operations which are easy to integrate given their concurrent processing and relative weighting within GDIP. For long-term autonomy and highly-safe operations, dealing with adverse conditions is crucial for autonomous vehicles, and this work pushes the boundaries of robust perception, getting a step closer to the ubiquity of autonomous vehicles.

\bibliography{root}

\begin{thebibliography}{10}
\providecommand{\url}[1]{#1}
\csname url@samestyle\endcsname
\providecommand{\newblock}{\relax}
\providecommand{\bibinfo}[2]{#2}
\providecommand{\BIBentrySTDinterwordspacing}{\spaceskip=0pt\relax}
\providecommand{\BIBentryALTinterwordstretchfactor}{4}
\providecommand{\BIBentryALTinterwordspacing}{\spaceskip=\fontdimen2\font plus
\BIBentryALTinterwordstretchfactor\fontdimen3\font minus
  \fontdimen4\font\relax}
\providecommand{\BIBforeignlanguage}[2]{{%
\expandafter\ifx\csname l@#1\endcsname\relax
\typeout{** WARNING: IEEEtran.bst: No hyphenation pattern has been}%
\typeout{** loaded for the language `#1'. Using the pattern for}%
\typeout{** the default language instead.}%
\else
\language=\csname l@#1\endcsname
\fi
#2}}
\providecommand{\BIBdecl}{\relax}
\BIBdecl

\bibitem{liu2022imageadaptive}
W.~Liu, G.~Ren, R.~Yu, S.~Guo, J.~Zhu, and L.~Zhang, ``Image-adaptive yolo for
  object detection in adverse weather conditions,'' in \emph{Proceedings of the
  AAAI Conference on Artificial Intelligence}, 2022.

\bibitem{chen2018domain}
Y.~Chen, W.~Li, C.~Sakaridis, D.~Dai, and L.~Van~Gool, ``Domain adaptive faster
  r-cnn for object detection in the wild,'' in \emph{Proceedings of the IEEE
  conference on computer vision and pattern recognition}, 2018, pp. 3339--3348.

\bibitem{Hsu_2020_WACV}
H.-K. Hsu, C.-H. Yao, Y.-H. Tsai, W.-C. Hung, H.-Y. Tseng, M.~Singh, and M.-H.
  Yang, ``Progressive domain adaptation for object detection,'' in
  \emph{Proceedings of the IEEE/CVF Winter Conference on Applications of
  Computer Vision (WACV)}, March 2020.

\bibitem{Tian_2021_ICCV}
K.~Tian, C.~Zhang, Y.~Wang, S.~Xiang, and C.~Pan, ``Knowledge mining and
  transferring for domain adaptive object detection,'' in \emph{Proceedings of
  the IEEE/CVF International Conference on Computer Vision (ICCV)}, October
  2021, pp. 9133--9142.

\bibitem{wu2021vector}
A.~Wu, R.~Liu, Y.~Han, L.~Zhu, and Y.~Yang, ``Vector-decomposed disentanglement
  for domain-invariant object detection,'' in \emph{Proceedings of the IEEE/CVF
  International Conference on Computer Vision}, 2021, pp. 9342--9351.

\bibitem{wu2021instance}
A.~Wu, Y.~Han, L.~Zhu, and Y.~Yang, ``Instance-invariant domain adaptive object
  detection via progressive disentanglement,'' \emph{IEEE Transactions on
  Pattern Analysis and Machine Intelligence}, 2021.

\bibitem{lin2021domain}
C.~Lin, Z.~Yuan, S.~Zhao, P.~Sun, C.~Wang, and J.~Cai, ``Domain-invariant
  disentangled network for generalizable object detection,'' in
  \emph{Proceedings of the IEEE/CVF International Conference on Computer
  Vision}, 2021, pp. 8771--8780.

\bibitem{gu2021pit}
Q.~Gu, Q.~Zhou, M.~Xu, Z.~Feng, G.~Cheng, X.~Lu, J.~Shi, and L.~Ma, ``Pit:
  Position-invariant transform for cross-fov domain adaptation,'' in
  \emph{Proceedings of the IEEE/CVF International Conference on Computer
  Vision}, 2021, pp. 8761--8770.

\bibitem{he2020domain}
Z.~He and L.~Zhang, ``Domain adaptive object detection via asymmetric tri-way
  faster-rcnn,'' in \emph{European conference on computer vision}.\hskip 1em
  plus 0.5em minus 0.4em\relax Springer, 2020, pp. 309--324.

\bibitem{sindagi2020prior}
V.~A. Sindagi, P.~Oza, R.~Yasarla, and V.~M. Patel, ``Prior-based domain
  adaptive object detection for hazy and rainy conditions,'' in \emph{European
  Conference on Computer Vision}.\hskip 1em plus 0.5em minus 0.4em\relax
  Springer, 2020, pp. 763--780.

\bibitem{zhang2020unified}
Z.~Zhang, L.~Zhao, Y.~Liu, S.~Zhang, and J.~Yang, ``Unified density-aware image
  dehazing and object detection in real-world hazy scenes,'' in
  \emph{Proceedings of the Asian Conference on Computer Vision}, 2020.

\bibitem{zhang2021domain}
S.~Zhang, H.~Tuo, J.~Hu, and Z.~Jing, ``Domain adaptive yolo for one-stage
  cross-domain detection,'' in \emph{Asian Conference on Machine
  Learning}.\hskip 1em plus 0.5em minus 0.4em\relax PMLR, 2021, pp. 785--797.

\bibitem{zhuang2020ifan}
C.~Zhuang, X.~Han, W.~Huang, and M.~Scott, ``ifan: Image-instance full
  alignment networks for adaptive object detection,'' in \emph{Proceedings of
  the AAAI Conference on Artificial Intelligence}, vol.~34, no.~07, 2020, pp.
  13\,122--13\,129.

\bibitem{sakaridis2020map}
C.~Sakaridis, D.~Dai, and L.~Van~Gool, ``Map-guided curriculum domain
  adaptation and uncertainty-aware evaluation for semantic nighttime image
  segmentation,'' \emph{IEEE Transactions on Pattern Analysis and Machine
  Intelligence}, 2020.

\bibitem{lengyel2021zero}
A.~Lengyel, S.~Garg, M.~Milford, and J.~C. van Gemert, ``Zero-shot day-night
  domain adaptation with a physics prior,'' in \emph{Proceedings of the
  IEEE/CVF International Conference on Computer Vision}, 2021, pp. 4399--4409.

\bibitem{Zheng_2022_WACV}
S.~Zheng and G.~Gupta, ``Semantic-guided zero-shot learning for low-light
  image/video enhancement,'' in \emph{Proceedings of the IEEE/CVF Winter
  Conference on Applications of Computer Vision (WACV) Workshops}, January
  2022, pp. 581--590.

\bibitem{huang2020dsnet}
S.-C. Huang, T.-H. Le, and D.-W. Jaw, ``Dsnet: Joint semantic learning for
  object detection in inclement weather conditions,'' \emph{IEEE transactions
  on pattern analysis and machine intelligence}, vol.~43, no.~8, pp.
  2623--2633, 2020.

\bibitem{liu2019griddehazenet}
X.~Liu, Y.~Ma, Z.~Shi, and J.~Chen, ``Griddehazenet: Attention-based
  multi-scale network for image dehazing,'' in \emph{Proceedings of the
  IEEE/CVF international conference on computer vision}, 2019, pp. 7314--7323.

\bibitem{guo2020zero}
C.~Guo, C.~Li, J.~Guo, C.~C. Loy, J.~Hou, S.~Kwong, and R.~Cong,
  ``Zero-reference deep curve estimation for low-light image enhancement,'' in
  \emph{Proceedings of the IEEE/CVF Conference on Computer Vision and Pattern
  Recognition}, 2020, pp. 1780--1789.

\bibitem{hu2018exposure}
Y.~Hu, H.~He, C.~Xu, B.~Wang, and S.~Lin, ``Exposure: A white-box photo
  post-processing framework,'' \emph{ACM Transactions on Graphics (TOG)},
  vol.~37, no.~2, pp. 1--17, 2018.

\bibitem{Yang_2020_CVPR}
W.~Yang, S.~Wang, Y.~Fang, Y.~Wang, and J.~Liu, ``From fidelity to perceptual
  quality: A semi-supervised approach for low-light image enhancement,'' in
  \emph{Proceedings of the IEEE/CVF Conference on Computer Vision and Pattern
  Recognition (CVPR)}, June 2020.

\bibitem{LI2021106617}
\BIBentryALTinterwordspacing
G.~Li, Y.~Yang, X.~Qu, D.~Cao, and K.~Li, ``A deep learning based image
  enhancement approach for autonomous driving at night,'' \emph{Knowledge-Based
  Systems}, vol. 213, p. 106617, 2021. [Online]. Available:
  \url{https://www.sciencedirect.com/science/article/pii/S0950705120307462}
\BIBentrySTDinterwordspacing

\bibitem{Zhang_2020_ACCV}
Z.~Zhang, L.~Zhao, Y.~Liu, S.~Zhang, and J.~Yang, ``Unified density-aware image
  dehazing and object detection in real-world hazy scenes,'' in
  \emph{Proceedings of the Asian Conference on Computer Vision (ACCV)},
  November 2020.

\bibitem{RTTS}
B.~Li, W.~Ren, D.~Fu, D.~Tao, D.~Feng, W.~Zeng, and Z.~Wang, ``Benchmarking
  single image dehazing and beyond,'' \emph{IEEE Transactions on Image
  Processing}, vol.~PP, pp. 1--1, 08 2018.

\bibitem{Exdark}
Y.~P. Loh and C.~S. Chan, ``Getting to know low-light images with the
  exclusively dark dataset,'' \emph{Computer Vision and Image Understanding},
  vol. 178, pp. 30--42, 2019.

\bibitem{ren2015faster}
S.~Ren, K.~He, R.~Girshick, and J.~Sun, ``Faster r-cnn: Towards real-time
  object detection with region proposal networks,'' \emph{Advances in neural
  information processing systems}, vol.~28, 2015.

\bibitem{he2017mask}
K.~He, G.~Gkioxari, P.~Doll{\'a}r, and R.~Girshick, ``Mask r-cnn,'' in
  \emph{Proceedings of the IEEE international conference on computer vision},
  2017, pp. 2961--2969.

\bibitem{redmon2018yolov3}
J.~Redmon and A.~Farhadi, ``Yolov3: An incremental improvement,'' \emph{arXiv
  preprint arXiv:1804.02767}, 2018.

\bibitem{lin2017focal}
T.-Y. Lin, P.~Goyal, R.~Girshick, K.~He, and P.~Doll{\'a}r, ``Focal loss for
  dense object detection,'' in \emph{Proceedings of the IEEE international
  conference on computer vision}, 2017, pp. 2980--2988.

\bibitem{liu2016ssd}
W.~Liu, D.~Anguelov, D.~Erhan, C.~Szegedy, S.~Reed, C.-Y. Fu, and A.~C. Berg,
  ``Ssd: Single shot multibox detector,'' in \emph{European conference on
  computer vision}.\hskip 1em plus 0.5em minus 0.4em\relax Springer, 2016, pp.
  21--37.

\bibitem{tian2019fcos}
Z.~Tian, C.~Shen, H.~Chen, and T.~He, ``Fcos: Fully convolutional one-stage
  object detection,'' in \emph{Proceedings of the IEEE/CVF international
  conference on computer vision}, 2019, pp. 9627--9636.

\bibitem{hnewa2021multiscale}
M.~Hnewa and H.~Radha, ``Multiscale domain adaptive yolo for cross-domain
  object detection,'' in \emph{2021 IEEE International Conference on Image
  Processing (ICIP)}.\hskip 1em plus 0.5em minus 0.4em\relax IEEE, 2021, pp.
  3323--3327.

\bibitem{li2017aod}
B.~Li, X.~Peng, Z.~Wang, J.~Xu, and D.~Feng, ``Aod-net: All-in-one dehazing
  network,'' in \emph{Proceedings of the IEEE international conference on
  computer vision}, 2017, pp. 4770--4778.

\bibitem{dong2020multi}
H.~Dong, J.~Pan, L.~Xiang, Z.~Hu, X.~Zhang, F.~Wang, and M.-H. Yang,
  ``Multi-scale boosted dehazing network with dense feature fusion,'' in
  \emph{Proceedings of the IEEE/CVF conference on computer vision and pattern
  recognition}, 2020, pp. 2157--2167.

\bibitem{he2010single}
K.~He, J.~Sun, and X.~Tang, ``Single image haze removal using dark channel
  prior,'' \emph{IEEE transactions on pattern analysis and machine
  intelligence}, vol.~33, no.~12, pp. 2341--2353, 2010.

\bibitem{zeng2020learning}
H.~Zeng, J.~Cai, L.~Li, Z.~Cao, and L.~Zhang, ``Learning image-adaptive 3d
  lookup tables for high performance photo enhancement in real-time,''
  \emph{IEEE Transactions on Pattern Analysis and Machine Intelligence}, 2020.

\bibitem{redmon2016you}
J.~Redmon, S.~Divvala, R.~Girshick, and A.~Farhadi, ``You only look once:
  Unified, real-time object detection,'' in \emph{Proceedings of the IEEE
  conference on computer vision and pattern recognition}, 2016, pp. 779--788.

\bibitem{pascal-voc-2007}
M.~Everingham, L.~Van~Gool, C.~K.~I. Williams, J.~Winn, and A.~Zisserman, ``The
  {PASCAL} {V}isual {O}bject {C}lasses {C}hallenge 2007 {(VOC2007)}
  {R}esults,''
  http://www.pascal-network.org/challenges/VOC/voc2007/workshop/index.html.

\bibitem{pascal-voc-2012}
------, ``The {PASCAL} {V}isual {O}bject {C}lasses {C}hallenge 2012 {(VOC2012)}
  {R}esults,''
  http://www.pascal-network.org/challenges/VOC/voc2012/workshop/index.html.

\bibitem{AtmospshereScatteringModel}
S.~Narasimhan and S.~Nayar, ``Vision and the atmosphere,'' \emph{International
  Journal of Computer Vision}, vol.~48, pp. 233--254, 07 2002.

\end{thebibliography}
\bibliographystyle{IEEEtran}
\end{document}